\documentclass{article}

 \PassOptionsToPackage{numbers, compress}{natbib}
\usepackage[final]{nips_2017} 
\usepackage{amsmath}

\usepackage[utf8]{inputenc} 
\usepackage[T1]{fontenc}    
\usepackage{hyperref}       
\usepackage{url}            
\usepackage{booktabs}       
\usepackage{amsfonts}       
\usepackage{nicefrac}       
\usepackage{microtype}      
\usepackage{graphicx}
\usepackage[font=small,labelfont=bf]{caption}

\title{Deep Learning Improves Template Matching by Normalized Cross Correlation}
\author{
  Davit Buniatyan, Thomas Macrina, Dodam Ih, Jonathan Zung, Sebastian Seung \\
  Neuroscience Institute and Computer Science Dept.\\
  Princeton University, Princeton, NJ 08544\\
  \texttt{\{davit, tmacrina, dih, jzung, sseung\}@princeton.edu} \\
}

\begin{document}

\maketitle

\begin{abstract}
Template matching by normalized cross correlation (NCC) is widely used for finding image correspondences. We improve the robustness of this algorithm by preprocessing images with "siamese" convolutional networks trained to maximize the contrast between NCC values of true and false matches. The improvement is quantified using patches of brain images from serial section electron microscopy. Relative to a parameter-tuned bandpass filter, siamese convolutional networks significantly reduce false matches. Furthermore, all false matches can be eliminated by removing a tiny fraction of all matches based on NCC values. The improved accuracy of our method could be essential for connectomics, because emerging petascale datasets may require billions of template matches to assemble 2D images of serial sections into a 3D image stack. Our method is also expected to generalize to many other computer vision applications that use NCC template matching to find image correspondences.
\end{abstract}

\section{Introduction}

Template matching by normalized cross correlation (NCC) is widely used in computer vision applications such as image registration, stereo matching, motion estimation, object detection and localization, and visual tracking \cite{, avants2008symmetric,  heo2011robust, lewis1995fast, luo2010fast, smeulders2014visual}. Here we show that the robustness of the algorithm can be improved by applying deep learning. Namely, if the template and source images are preprocessed by a convolutional network, the rate of false matches can be significantly reduced, and NCC output becomes more useful for rejecting suspect matches. The training of the convolutional network follows the "siamese network" method of learning a measure of similarity from pairs of input images \cite{chopra2005learning}. The learning is only weakly supervised, in the sense that a true match to the template should exist somewhere in the source image, but the location of that match is not an input to the learning procedure.  If NCC already works fairly well for template matching with raw images, then the incorporation of deep learning is expected to improve its accuracy further.

We test the power of our technique using images acquired by serial section electron microscopy (EM). NCC template matching is commonly applied to image patches in the course of assembling a 3D image stack from 2D images of individual sections \cite{preibisch2009bead, saalfeld2012elastic}. Achieving highly precise alignment between successive sections is critical for the accuracy of the subsequent step of tracing fine neurites through the image volume. Erroneous matches may arise because sections are deformed, distorted, or damaged during collection, and defects may also arise during imaging \cite{saalfeld2012elastic}. An image of a (0.1 mm)$^3$ brain volume is roughly a teravoxel \cite{lichtman2014bigdata}, and a high quality assembly could require up to 100 million template matches \cite{saalfeld2012elastic}. Every false match leads to tracing errors in multiple neurites, so even a small error rate across so many matches can have devastating consequences.


In empirical tests, we find that the error rate of template matching on raw serial EM images is on the order of 1 to 3\%. Preprocessing with a bandpass filter lowers the error rate \cite{berg2001geometric}, and substituting convolutional networks improves upon that error rate by a factor of 2-7x. The overall result is an error rate of 0.05 to 0.30\%. A common strategy for reducing false matches is to reject those that are suspect according to some criteria \cite{saalfeld2012elastic}. This can be problematic if too many true matches are rejected and there are not enough matches to describe the deformation in a given region, which can also lead to tracing errors in multiple neurites. We show that NCC output provides superior rejection efficiency once deep learning is incorporated. To achieve zero false matches under our most accurate conditions, we need only reject 0.12\% of the true matches based on NCC output, an improvement of 3.5x over the efficiency based on a bandpass filter.

The idea of using deep learning to improve NCC template matching for image correspondences is simple and obvious, but has been little explored as far as we know. The closest antecedent of our work introduced an NCC layer inside a network used for the person identification problem \cite{subramaniam2016deep}. Recent work applying deep learning to image correspondences avoids template matching and instead trains a convolutional network to directly output a vector field \cite{dosovitskiy2015flownet, long2014convnets, pathak2016learning}. This approach is well-suited for computing dense correspondences, while template matching makes sense for computing sparse sets of corresponding points.

An advantage of the template matching approach is its interpretability. The height of an NCC peak provides information about the goodness of a match, and the width of an NCC peak provides information about the accuracy of spatial localization. Furthermore, one can examine the convolutional network output to see what image features are being used to compute matches. In our application to serial EM images, it appears that the network detects mitochondria. It learns to suppress image defects that arise from brightness-contrast fluctuations and damaged sections. It also learns to suppress high contrast edges of blood vessels. When such edges are present, they tend to produce a strong NCC peak, but the peak is very wide due to the near straightness of the edges, leading to imprecise spatial localization.

\begin{figure}[h]
  \centering
  
  \includegraphics[width=\linewidth]{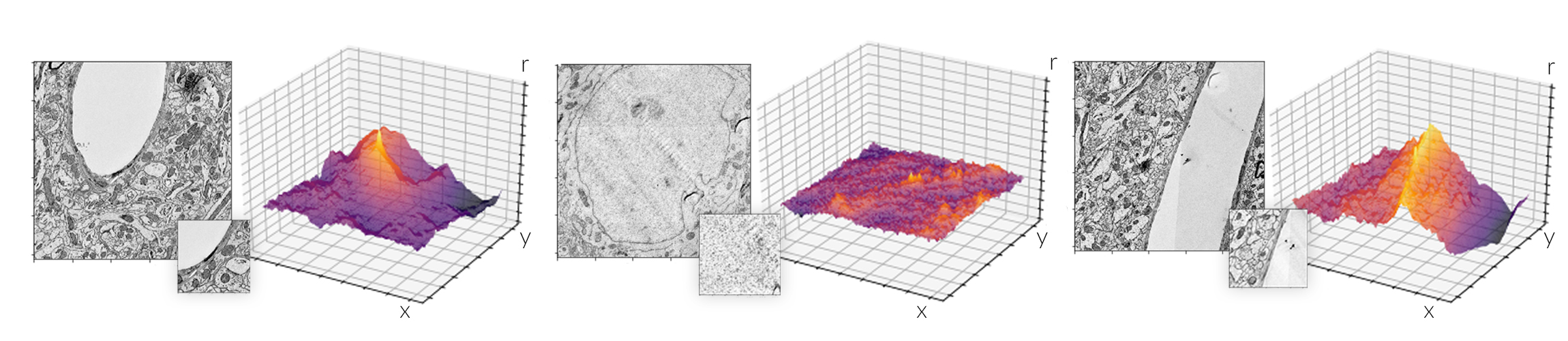}
  \caption{Three examples of template matching by 3D NCC correlograms. The large image in each example is the source image, and the small inset is the template. The correlogram plots the Pearson correlation coefficient $r$ values for all possible translations of the template across the source. The first image shows a correlogram with a well-localized maximum ("peak"). The second and third images show correlograms that fail to produce a clearly distinct maximum.}
  \label{ncc_examples}
\end{figure}

\section{Methods}

\subsection{Weakly Supervised Similarity Metric Learning by Siamese Convolutional Nets}

The inputs to the NCC are a template image and a larger source image. If the template image is placed somewhere inside the borders of the source image, the template pixels are in one-to-one correspondence with a subset of the source pixels, and the Pearson correlation coefficient can be computed for the pixel pairs. This computation can be done for all placements of the template image inside the source image, yielding an output image called the normalized cross-correlogram or correlogram for short (see Fig. \ref{ncc_examples}). The location in the correlogram with the largest Pearson coefficient is considered the location at which the template matches the source. 

Ideally, the correlogram should have a high and narrow peak only at the location of a true match, and should be low at all other locations. There should be no peak at all if there is no good match between the template and source. In practice, there can be peaks at spurious locations, leading to false matches. Another failure mode is a wide peak near a true match, leading to imprecise spatial localization.

To reduce the failure rate, one could apply preprocessing to the template and source images prior to computing the NCC. The preprocessing step can be trained from data using standard methods for supervised learning of a similarity metric \cite{kulis2013metric, yang2006distance}. Given pairs of points in a space $X$ that are known to be similar or dissimilar, and a similarity measure $S:\mathbb{R}^n\times\mathbb{R}^n\to \mathbb{R}$, the method is to learn an embedding $\psi:X\rightarrow \mathbb{R}^n$ such that $S(\psi(x),\psi(y))$ is large for similar $(x,y)$ and small for dissimilar $(x,y)$. If the embedding function $\psi$ is a neural network, then the technique is known as "siamese networks"\cite{bromley1993signature} because identical networks are applied to both $x$ and $y$ \cite{chopra2005learning}. 





\begin{figure}[h]
  \centering
  \includegraphics[width=\linewidth]{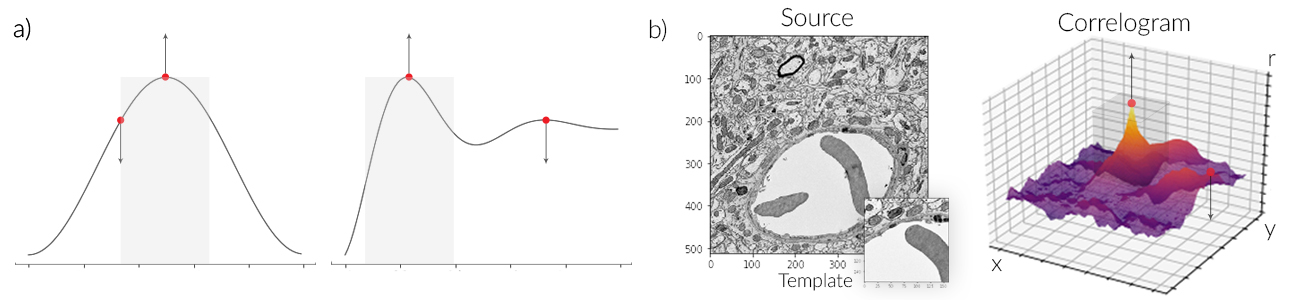}
  \caption{Loss function intuition. (a) 2D toy example. Left, make a correlogram with a wide peak more narrow. Right, promote the first peak and diminish the second peak. (b) Real 3D example. Generate NCC correlogram from template and source, then promote the first peak and diminish the second peak.}
  \label{loss_function_2D}
\end{figure}

We train siamese convolutional networks by repeating the following for template-source pairs that are known to contain a true match:
\begin{enumerate}
\item Compute the correlogram for source and template image. 
\item Find the peak of the correlogram. 
\item Make a gradient update to the convolutional net that increases the height of the peak. 
\item Draw a small box around the peak of the correlogram.
\item Find the maximum of the correlogram outside the box, and call this the "secondary peak."
\item Make a gradient update to the convolutional net that decreases the secondary peak.
\end{enumerate}
The cost function for the above algorithm is the difference in the heights of the primary and secondary peaks, which we will call the "correlation gap." The cost function has two purposes, depending on the shape of the correlogram (Fig. \ref{loss_function_2D}). If the primary peak is wider than the box, then the secondary peak will not actually be a local maximum (Fig. \ref{loss_function_2D}). In this case, the cost function encourages narrowing of the primary peak, which is good for precise spatial localization. The size of the box in the algorithm represents the desired localization accuracy. In other cases, the secondary peak will be a true local maximum, in which case the purpose of the cost function is to suppress peaks corresponding to false matches. 

The above algorithm corresponds to similarity metric learning if the primary peak indeed represents a true match and the secondary peak represents a false match. In fact, the NCC does have a nonzero error rate, which means that some of the examples in the training have incorrect labeling. However, if the error rate starts out small, one can hope that the similarity metric learning will make it even smaller. Our algorithm requires supervision, in the sense that a good match should exist between each source-template pair.  However, the location of the match is not required as an input to the learning, so the supervision is fairly weak.

By itself, the above algorithm may lead to pathological solutions in which the network is able to minimize the cost function by ignoring the input image. To avoid these solutions, one can additionally train on source-template pairs that are known to contain no good match. Since these are dissimilar pairs, the goal of learning is to reduce the peak of the NCC.
\begin{enumerate}
\item Compute the correlogram for source and template image. 
\item Find the peak of the correlogram. 
\item Make a gradient update to the convolutional net that decreases the height of the peak. 
\end{enumerate}
Dissimilar pairs can be artificially generated by permuting the source and template images within a batch. 


\label{gen_inst}

\begin{figure}[h]
  \centering
  
  \includegraphics[width=\linewidth]{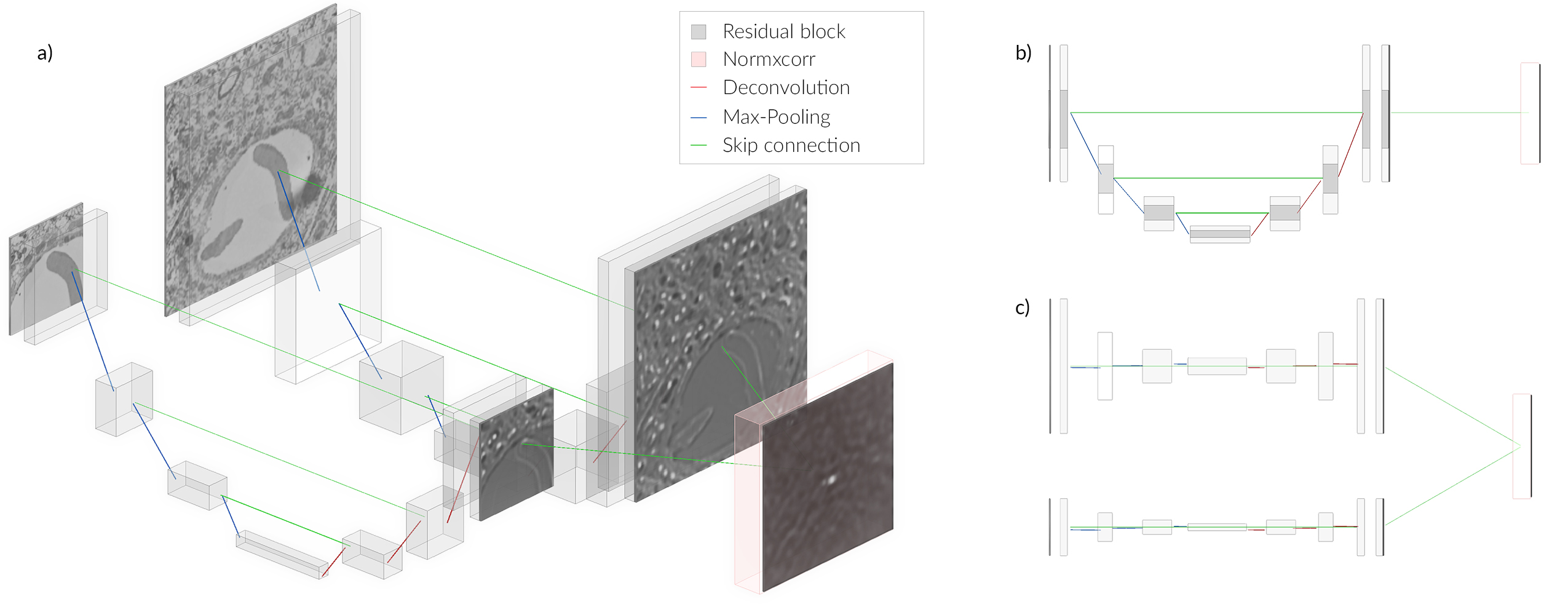}
  \caption{Architecture diagram of two channel neural network with NCC block as a bridge. a) Perspective view of the network. Gray boxes represent residual blocks that are either connected to each other through skip connections, max-pooling and convolution layers. b) Left view where U-Net architecture can be easily seen. c) Top view shows two Siamese networks with weight-sharing.}
  \label{architecture}
\end{figure}

\subsection{Implementation} 
  
 Fig. \ref{architecture} depicts siamese convolution networks, i.e., two networks with the same architecture and weight-sharing between the networks. The architecture is FusionNet \citet{hegde2016fusionnet}, which is a variant of U-Net. Instead of convolution blocks it uses residual blocks consisting of three convolution layers and a skip connection from the first layer to the last. Instead of concatenation at each level, the output of the left-side is summed with the right-side. This network also enforces symmetric input-output resolution.
 
 The input size of both networks plays a crucial role because it defines the sparseness of features that the network will preserve for optimizing the NCC. The template and the source image are both squares with sizes 160px and 512px. We consistently use $3\times 3$ convolution layers with $tanh$ non-linearity. At each level the number of features is doubled starting with 8 up to 64 channels. The output of FusionNet is passed through another convolution layer that feeds to the NCC layer. 

 The NCC layer is implemented in TensorFlow using FFT \citet{lewis1995fast} and can handle batches and multiple channels effectively. The loss layer takes as input the NCC correlogram, computes the maximum peak, removes a 20px square window centered at the peak, then computes the next maximum value that represents the second peak. The initial framework for the loss is defined to maximize the difference between the first and second peaks during training. 
 

 Training alternated between a batch of eight source-template pairs and then the same batch with randomly permuted source-template pairings. Gradient descent used the Adam optimizer with learning rate of 0.0005. Training converged within 10,000 iterations. The training data consisted of pairs of inputs sampled from an affine-aligned stack of images that contained non-affine deformations. It is recommended either to choose a dataset with enhanced pathological cases that the network is expected to handle or to use data augmentation for covering the problem space of possible damages and deformations. During the training we randomly cropped the source and template images such that the position of the peak is randomly distributed. Also, to increase the size of the training dataset we used random rotations of both inputs by 90, 180, 270 degrees.


\section{Experiments}

We validated our model on 95 serial images from the training set, an unpublished EM dataset with a resolution of 7x7x40nm$^3$. Each image was 15,000x15,000px, and had been roughly aligned with an affine model but still contained considerable non-affine distortions up to 250px (full resolution).

From the serial images, we produced three datasets: raw images (\textit{raw}), images preprocessed with a circular Gaussian bandpass filter that was optimally tuned to produce a low number of false matches (\textit{bandpass}), and images preprocessed with our convolutional net by applying the larger convolutional channel across the entire image, upsampling and blending accordingly (\textit{convnet}). We then varied the parameters of our template matching procedure, varying the template image size between small and large (\textit{160px} and \textit{224px}), and matching between neighboring images (\textit{adjacent}) as well as the next-nearest neighbors (\textit{across}). For matching between next-nearest neighbors, a slightly different bandpass parameter was used, as the optimal filter for next-nearest neighbors differed from the filter for neighboring images. The network used was identical in all experiments, having been trained on 160px template and 512px source patches from adjacent sections. Table \ref{table:parameters} summarizes the training and experiment parameters.

\begin{table}[h]
    \caption{Image parameters for training and testing. Unless otherwise noted, resolutions are given after 3x downsampling where 1px represents 21nm.}
    \centering
     \small
    \begin{tabular}{l*{6}{c}r}
    
        \toprule
        & \multicolumn{1}{c}{Training} & \multicolumn{2}{c}{Adjacent}   &  \multicolumn{2}{c}{Across}\\
        \cmidrule{2-6}
        
        Template size   & 160px & 160px & 224px & 160px & 224px \\
        Source size     & 512px &\multicolumn{2}{c}{512px} &\multicolumn{2}{c}{512px} \\
        Section depth   & 40nm &\multicolumn{2}{c}{40nm} &\multicolumn{2}{c}{80nm} \\
        \cmidrule{2-6}
        Section size (full res.)    & ~33,000px & \multicolumn{2}{c}{15,000px} & \multicolumn{2}{c}{15,000px} \\
        No. of sections & 1040 & \multicolumn{2}{c}{95} & \multicolumn{2}{c}{48}\\
        No. of matches & 10,000 & \multicolumn{2}{c}{~144,000} & \multicolumn{2}{c}{~72,000}\\
        \cmidrule{2-6}
        Bandpass $\sigma$ (full res.) & N/A & \multicolumn{2}{c}{2.0-12.0px} & \multicolumn{2}{c}{2.5-25.0px}\\
       
        \bottomrule
    \end{tabular}
\label{table:parameters}
\end{table}

\begin{figure}[h]
  \centering
  
  \includegraphics[width=\linewidth]{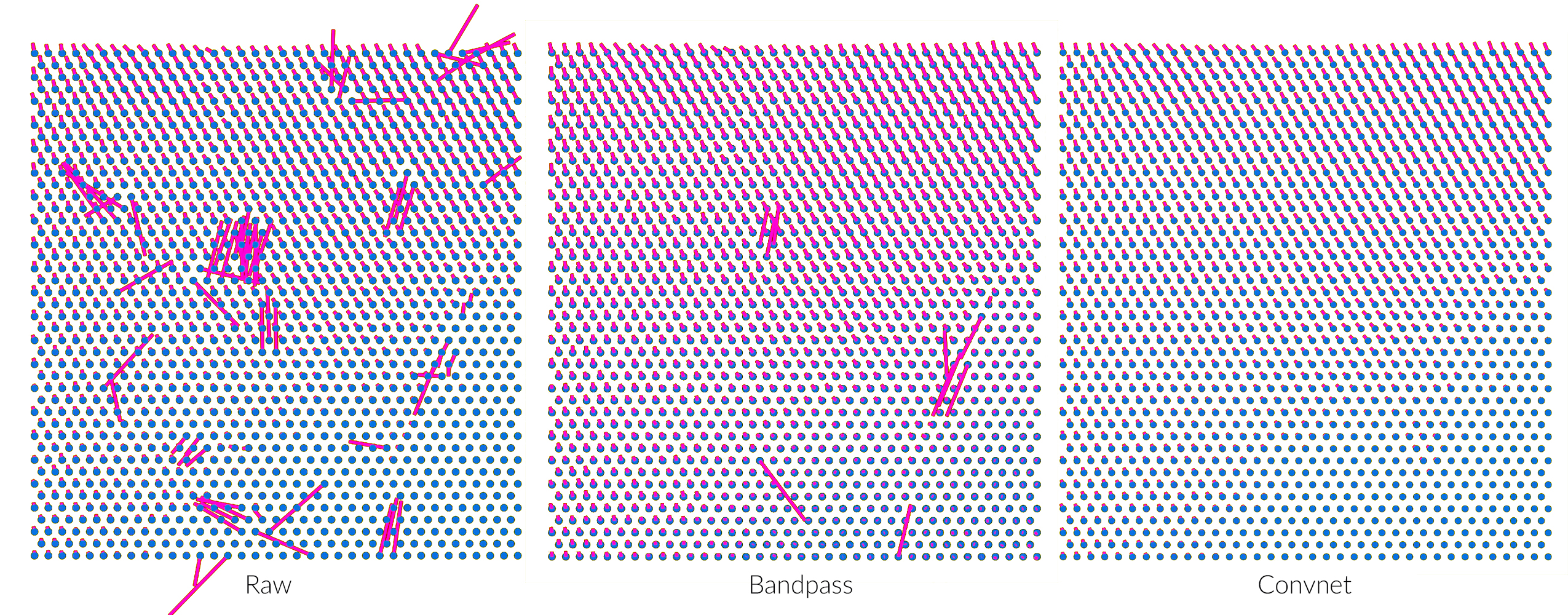} 
  \caption{Example displacement vector fields for each image condition. Representation of output of template matching in a regular triangular grid (edge length 400px at full resolution) across a 15,000x15,000px image. Each node represents the centerpoint of a template image used in the template matching procedure. Each vector represents the displacement of that template image to its matching location in its source image. Matches shown are based on 224px template size on across (next-nearest neighbor) sections. Raw: matches on the raw images. Bandpass: matches on images filtered with a Gaussian bandpass filter; Convnet: matches on the output of the convolutional network processed image.}
  \label{vectorfield}
\end{figure}

In each experiment, both the template and the source images were downsampled by a factor of 3 before NCC, so that 160px and 224px templates were 480px and 672px at full resolution, while the source image was fixed at 512px downsampled (1,536px full resolution). The template matches were taken in a triangular grid covering the image, with an edge length of 400px at full resolution (Fig. \ref{vectorfield} shows the locations of template matches across an image).

\begin{figure}[h]
  \centering
  \includegraphics[width=250pt]{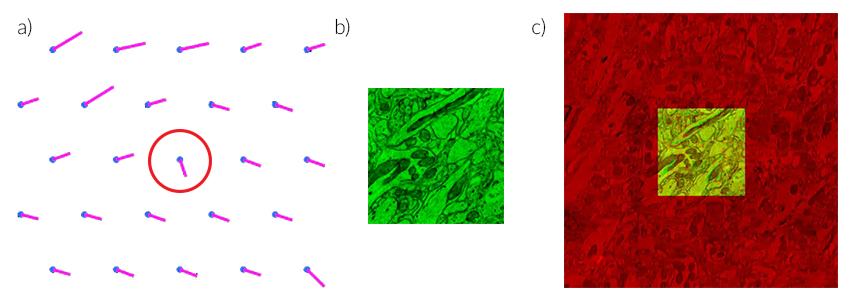}
  \caption{Manual inspection difficulties. a) The vector field around a match (circled in red) that prominently differs from its neighbors. b) The template for the match, showing many neurites parallel to the sectioning plane. c) The false color overlay of the template (green) over the source image (red) at the matched location, establishing the match as true.}
  \label{difficult_match}
\end{figure}

Our first method to evaluate performance was to compare error rates. Errors were detected manually, using a tool that allowed human annotators to inspect the template matching inputs and outputs. The tool is based on the visualization of the displacement vectors that result from each template match across a section, as shown in Fig. \ref{vectorfield}. Any match that significantly differed (over 50px) from its neighbors were rejected, and matches that differed from neighbors but not significantly were individually inspected for correctness by visualizing a false color overlay of the template over the source at the match location. The latter step was needed as there were many true matches that deviated prominently from its neighbors: the template patch could contain neurites or other features parallel to the sectioning plane, resulting in large motions of specific features in a random direction that may not be consistent with the movement of the larger area around the template (see Fig. \ref{difficult_match} for an example of this behavior). Table \ref{table:error_rates} summarizes the error counts in each experiment.

\begin{table}[h]
    \caption{False matches for each image condition across experiments. Total possible adjacent matches: 144,500. Total possible across matches: 72,306.}
    \centering
     \small
    \begin{tabular}{l*{9}{c}r}
    
        \toprule
        &  \multicolumn{4}{c}{Adjacent}   &  \multicolumn{4}{c}{Across}\\
        \cmidrule{2-9}
        Template size    & \multicolumn{2}{c}{160px}  & \multicolumn{2}{c}{224px} & \multicolumn{2}{c}{160px} & \multicolumn{2}{c}{224px} \\
        \hline
        Raw             & 1,778 & 1.23\% & 827 & 0.57\% & 2,105 & 2.91\% & 1,068 & 1.48\% \\
        Bandpass        & 480 & 0.33\% & 160 & 0.11\% & 1,504 & 2.08\% & 340 & 0.47\% \\
        Convnet         & \textbf{261} & \textbf{0.18\%} & \textbf{69} & \textbf{0.05\%} & \textbf{227} & \textbf{0.31\%} & \textbf{45} & \textbf{0.06\%} &  \\
        \bottomrule
    \end{tabular}
\label{table:error_rates}
\end{table}

To ensure that fewer false matches were not coming at the expense of true matches, we evaluated the overlap between true match sets created by the bandpass images and our convnet images. Table \ref{table:true_overlap} summarizes how many true matches were unique to the bandpass, convnet, or neither.

\begin{table}[h]
\caption{Dissociation of true matches set between the bandpass and convnet. Counts of true matches per category. Total possible adjacent matches: 144,500. Total possible across matches: 72,306.}
\centering
\small
    \begin{tabular}{l*{9}{c}r}
        \toprule
        &  \multicolumn{4}{c}{Adjacent}   &  \multicolumn{4}{c}{Across}\\
        \cmidrule{2-9}
        Template size    & \multicolumn{2}{c}{160px}  & \multicolumn{2}{c}{224px} & \multicolumn{2}{c}{160px} & \multicolumn{2}{c}{224px} \\
        \hline
        Neither         & 144 & 0.10\% & 54 & 0.04\% & 162 & 0.22\% & 33 & 0.05\% &  \\
        Bandpass only             & 117 & 0.08\% & 15 & 0.01\% & 65 & 0.09\% & 12 & 0.02\% \\
        Convnet only        & 336 & 0.23\% & 106 & 0.07\% & 1342 & 1.86\% & 307 & 0.42\% \\
        \bottomrule
    \end{tabular}

\label{table:true_overlap}
\end{table}

To assess how easily false matches could be removed, we evaluated matches with the following criteria:
\begin{itemize}
    \item \textit{norm}: The Euclidean norm of the displacement required to move the template image to its match location in the source image, at full resolution.
    \item \textit{r max}: The first peak of the correlogram serves as a proxy for confidence in the match.
    \item \textit{r delta}: The difference between the first peak and second peak (after removing a 5px square window surrounding the first peak) of the correlogram provides some estimate of the certainty there is no other likely match in the source image, and the criteria the convnet was trained to optimize.
\end{itemize}

\begin{figure}[h]
  \centering
  
  \includegraphics[width=\linewidth]{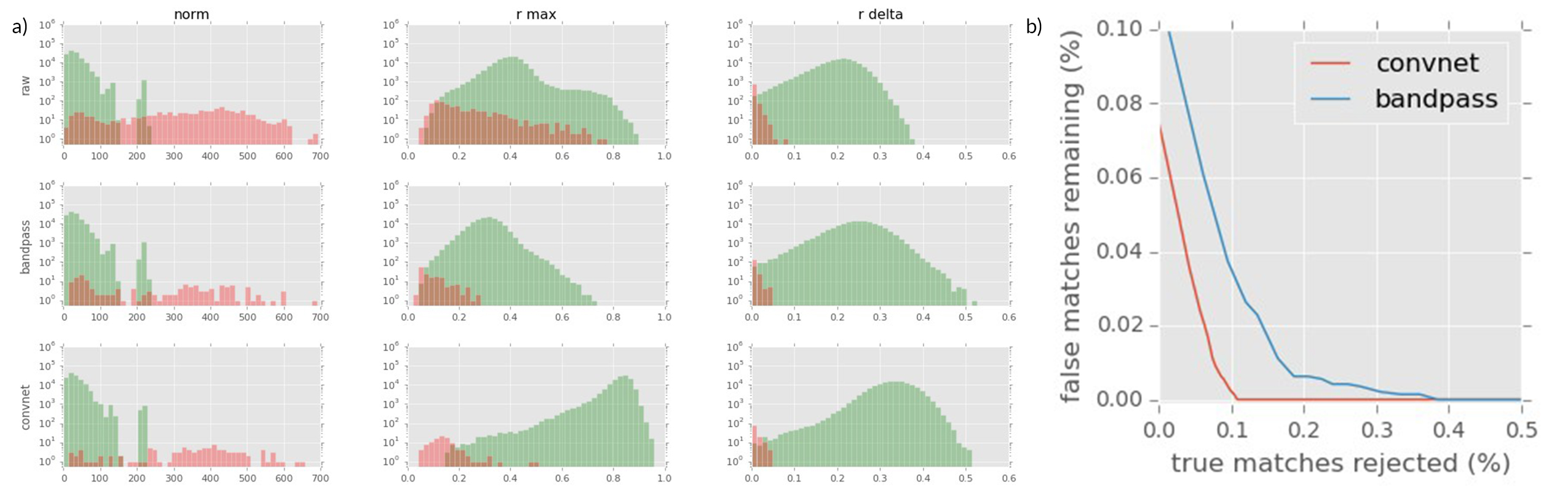}
  \caption{Match criteria for adjacent 224px experiment. (a) Distributions of the three template match criteria for each image condition. Red bars represent counts of false matches that were manually identified. Green bars represent counts of true matches. (b) Percentage of true matches that must be rejected to reduce the error rate when varying the \textit{r delta} criterion. See the Appendix for distributions \& rejection curves from other experiments.}
  \label{criteria_distributions}
\end{figure}

These criteria can serve as useful heuristics to accept or reject matches to approximate the unknown partitions for the true and erroneous matches. The less overlap between the actual distributions when projected onto the criterion dimension, the more useful that criterion. Fig. \ref{criteria_distributions} plots these three criteria across the three image conditions.

\section{Discussion}


In all experiments, the images preprocessed by our convnet consistently produced fewer false matches than the other two sets of images with a reduction factor of 2-7x (see Table \ref{table:error_rates}). The convnet produced matches in the vast majority of cases that the bandpass produced matches. It did introduce some false matches that the bandpass did not, but it correctly identified 3-20 times as many additional true matches relatively (see Table \ref{table:true_overlap}). The majority of the false matches in the convnet output were also present in the bandpass case, which establishes the convnet as superior to and not merely different from bandpass.

Notably, the reduction in error from applying the convnet generalized well to both harder (across sections at 160px and 224px template sizes) and easier (adjacent sections at 224px template size) tasks. In fact, the convnet provided larger gains in experiments other than on the task on which it was trained. This ability to generalize is crucial to applications as different template matching parameters are often needed at different stages of the alignment process. The results suggest that a single convnet may be used throughout the range of speed-accuracy tradeoffs (smaller-larger template size) as well as in dealing with missing sections (across).

\begin{figure}[h]
  \centering
  
  \includegraphics[width=\linewidth]{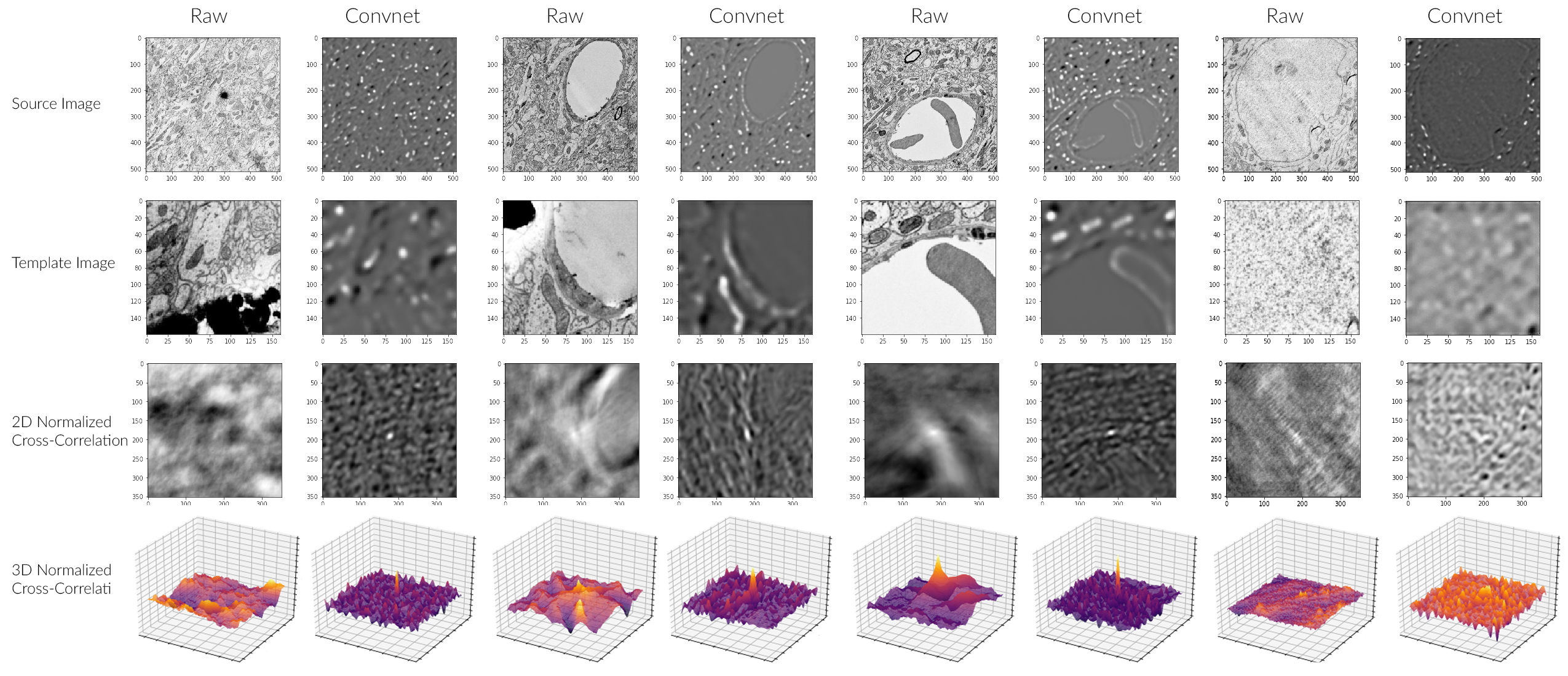}
  \caption{Difficult examples from the dataset with damaged areas \& local brightness changes. Correlograms are projected to be 2D with white pixels having higher \textit{r} values. The last column is an example of failure by the convnet.}
  \label{ncc_examples2}
\end{figure}

Inspecting the filtered image, the convnet seems to identify keypoints (small dark objects that localize well, such as mitochondria) and suppress objects that do not localize well (e.g. lines, such as cell membranes, or consistently patterned regions, such as regions inside cell bodies and blood vessels). See Fig. \ref{ncc_examples2} for examples from the convnet image set. The convnet fails when the template does not contain the keypoints it has learned to identify. The last column in Fig. \ref{ncc_examples2} contains a template that is almost completely occupied by a cell body, and the convnet failed to find the true match. The raw image can be more useful in those cases, because it can match on corner-like edges (see Sup. Fig. 8 in the Appendix). This can be improved by biasing the training set with more of these pathological examples.

Fortunately, when these false matches do occur with the convnet, we can reject them efficiently using our match criteria. The convnet transformed the true match distributions for \textit{r max} and \textit{r delta} to be more left-skewed, while the erroneous match distribution for \textit{r delta} remain with lower values (see Fig. \ref{criteria_distributions}a), resulting in a distribution more amenable to accurate error rejection. For the case of adjacent sections with 224px templates, we can remove every error in our convnet output by rejecting matches with an \textit{r delta} below 0.05, which removes only 0.12\% of the true matches. The same threshold also removes all false matches in the bandpass outputs, but removes 0.40\% of the true matches (see Fig. \ref{criteria_distributions}b). This 3.5x improvement in rejection efficiency is critical to balancing the trade-off between complete elimination of false matches and retaining as many true matches as possible.

The improvement in rejection efficiency also generalized well across experiments, as evident in the Appendix, Sup. Fig. 15. Achieving a 0.1\% error rate on the most difficult task we tested (across, 160px template size) required rejecting 20\% of the true matches on bandpass, while less than 1\% rejection of true matches was sufficient with the convnet.







\section{Conclusions}

Combining NCC with deep learning reduces false matches from template matching. It also improves the efficiency by which those false matches can be removed so that a minimal number of true matches are rejected. This is a very promising technique that offers us the ability to significantly increase the throughput of our alignment process while maintaining the precision we require. We expect this technique to serve well in other areas that demand such high-quality template matching.

We would like to explore how well this technique generalizes from one EM dataset to another, as well as to investigate if there is transfer learning that could benefit the segmentation convolutional network that follows the alignment process.

\medskip

\small

\bibliography{sources}

\begin{thebibliography}{18}
\providecommand{\natexlab}[1]{#1}
\providecommand{\url}[1]{\texttt{#1}}
\expandafter\ifx\csname urlstyle\endcsname\relax
  \providecommand{\doi}[1]{doi: #1}\else
  \providecommand{\doi}{doi: \begingroup \urlstyle{rm}\Url}\fi

\bibitem[Avants et~al.(2008)Avants, Epstein, Grossman, and
  Gee]{avants2008symmetric}
Brian~B Avants, Charles~L Epstein, Murray Grossman, and James~C Gee.
\newblock Symmetric diffeomorphic image registration with cross-correlation:
  evaluating automated labeling of elderly and neurodegenerative brain.
\newblock \emph{Medical image analysis}, 12\penalty0 (1):\penalty0 26--41,
  2008.

\bibitem[Berg and Malik(2001)]{berg2001geometric}
Alexander~C Berg and Jitendra Malik.
\newblock Geometric blur for template matching.
\newblock In \emph{Computer Vision and Pattern Recognition, 2001. CVPR 2001.
  Proceedings of the 2001 IEEE Computer Society Conference on}, volume~1, pages
  I--I. IEEE, 2001.

\bibitem[Bromley et~al.(1993)Bromley, Bentz, Bottou, Guyon, LeCun, Moore,
  S{\"a}ckinger, and Shah]{bromley1993signature}
Jane Bromley, James~W. Bentz, L{\'e}on Bottou, Isabelle Guyon, Yann LeCun,
  Cliff Moore, Eduard S{\"a}ckinger, and Roopak Shah.
\newblock Signature verification using a "siamese" time delay neural network.
\newblock \emph{IJPRAI}, 7\penalty0 (4):\penalty0 669--688, 1993.

\bibitem[Chopra et~al.(2005)Chopra, Hadsell, and LeCun]{chopra2005learning}
Sumit Chopra, Raia Hadsell, and Yann LeCun.
\newblock Learning a similarity metric discriminatively, with application to
  face verification.
\newblock In \emph{Computer Vision and Pattern Recognition, 2005. CVPR 2005.
  IEEE Computer Society Conference on}, volume~1, pages 539--546. IEEE, 2005.

\bibitem[Dosovitskiy et~al.(2015)Dosovitskiy, Fischer, Ilg, Hausser, Hazirbas,
  Golkov, van~der Smagt, Cremers, and Brox]{dosovitskiy2015flownet}
Alexey Dosovitskiy, Philipp Fischer, Eddy Ilg, Philip Hausser, Caner Hazirbas,
  Vladimir Golkov, Patrick van~der Smagt, Daniel Cremers, and Thomas Brox.
\newblock Flownet: Learning optical flow with convolutional networks.
\newblock In \emph{Proceedings of the IEEE International Conference on Computer
  Vision}, pages 2758--2766, 2015.

\bibitem[Hegde and Zadeh(2016)]{hegde2016fusionnet}
Vishakh Hegde and Reza Zadeh.
\newblock Fusionnet: 3d object classification using multiple data
  representations.
\newblock \emph{arXiv preprint arXiv:1607.05695}, 2016.

\bibitem[Heo et~al.(2011)Heo, Lee, and Lee]{heo2011robust}
Yong~Seok Heo, Kyong~Mu Lee, and Sang~Uk Lee.
\newblock Robust stereo matching using adaptive normalized cross-correlation.
\newblock \emph{IEEE Transactions on Pattern Analysis and Machine
  Intelligence}, 33\penalty0 (4):\penalty0 807--822, 2011.

\bibitem[Kulis et~al.(2013)]{kulis2013metric}
Brian Kulis et~al.
\newblock Metric learning: A survey.
\newblock \emph{Foundations and Trends{\textregistered} in Machine Learning},
  5\penalty0 (4):\penalty0 287--364, 2013.

\bibitem[Lewis(1995)]{lewis1995fast}
John~P Lewis.
\newblock Fast template matching.
\newblock In \emph{Vision interface}, volume~95, pages 15--19, 1995.

\bibitem[Lichtman et~al.(2014)Lichtman, Pfister, and
  Shavit]{lichtman2014bigdata}
Jeff~W Lichtman, Hanspeter Pfister, and Nir Shavit.
\newblock The big data challenges of connectomics.
\newblock \emph{Nat Neurosci}, 17\penalty0 (11):\penalty0 1448--1454, 2014.

\bibitem[Long et~al.(2014)Long, Zhang, and Darrell]{long2014convnets}
Jonathan~L Long, Ning Zhang, and Trevor Darrell.
\newblock Do convnets learn correspondence?
\newblock In \emph{Advances in Neural Information Processing Systems}, pages
  1601--1609, 2014.

\bibitem[Luo and Konofagou(2010)]{luo2010fast}
Jianwen Luo and Elisa~E Konofagou.
\newblock A fast normalized cross-correlation calculation method for motion
  estimation.
\newblock \emph{IEEE transactions on ultrasonics, ferroelectrics, and frequency
  control}, 57\penalty0 (6):\penalty0 1347--1357, 2010.

\bibitem[Pathak et~al.(2016)Pathak, Girshick, Doll{\'a}r, Darrell, and
  Hariharan]{pathak2016learning}
Deepak Pathak, Ross Girshick, Piotr Doll{\'a}r, Trevor Darrell, and Bharath
  Hariharan.
\newblock Learning features by watching objects move.
\newblock \emph{arXiv preprint arXiv:1612.06370}, 2016.

\bibitem[Preibisch et~al.(2009)Preibisch, Saalfeld, Rohlfing, and
  Tomancak]{preibisch2009bead}
Stephan Preibisch, Stephan Saalfeld, Torsten Rohlfing, and Pavel Tomancak.
\newblock Bead-based mosaicing of single plane illumination microscopy images
  using geometric local descriptor matching.
\newblock In \emph{SPIE Medical Imaging}, pages 72592S--72592S. International
  Society for Optics and Photonics, 2009.

\bibitem[Saalfeld et~al.(2012)Saalfeld, Fetter, Cardona, and
  Tomancak]{saalfeld2012elastic}
Stephan Saalfeld, Richard Fetter, Albert Cardona, and Pavel Tomancak.
\newblock Elastic volume reconstruction from series of ultra-thin microscopy
  sections.
\newblock \emph{Nature methods}, 9\penalty0 (7):\penalty0 717--720, 2012.

\bibitem[Smeulders et~al.(2014)Smeulders, Chu, Cucchiara, Calderara, Dehghan,
  and Shah]{smeulders2014visual}
Arnold~WM Smeulders, Dung~M Chu, Rita Cucchiara, Simone Calderara, Afshin
  Dehghan, and Mubarak Shah.
\newblock Visual tracking: An experimental survey.
\newblock \emph{IEEE Transactions on Pattern Analysis and Machine
  Intelligence}, 36\penalty0 (7):\penalty0 1442--1468, 2014.

\bibitem[Subramaniam et~al.(2016)Subramaniam, Chatterjee, and
  Mittal]{subramaniam2016deep}
Arulkumar Subramaniam, Moitreya Chatterjee, and Anurag Mittal.
\newblock Deep neural networks with inexact matching for person
  re-identification.
\newblock In \emph{Advances in Neural Information Processing Systems}, pages
  2667--2675, 2016.

\bibitem[Yang and Jin(2006)]{yang2006distance}
Liu Yang and Rong Jin.
\newblock Distance metric learning: A comprehensive survey.
\newblock \emph{Michigan State Universiy}, 2\penalty0 (2), 2006.

\end{thebibliography}
\bibliographystyle{plainnat}

\newpage

\end{document}